# Log-linear Error State Model Derivation without Approximation for INS[1]

Lubin Chang and Yarong Luo

*ABSTRACT* —Through assembling the navigation parameters as matrix Lie group state, the corresponding inertial navigation system (INS) kinematic model possesses a group-affine property. The Lie logarithm of the navigation state estimation error satisfies a log-linear autonomous differential equation. These log-linear models are still applicable even with arbitrarily large initial errors, which is very attractive for INS initial alignment. However, in existing works, the log-linear models are all derived based on first-order linearization approximation, which seemingly goes against their successful applications in INS initial alignment with large misalignments. In this work, it is shown that the log-linear models can also be derived without any approximation, the error dynamics for both left and right invariant error in continuous time are given in matrix Lie group $SE_2(3)$ for the first time. This work provides another evidence for the validity of the log-linear model in situations with arbitrarily large initial errors.

*INDEX TERMS*—Inertial navigation system, initial alignment, invariant extended Kalman filter

## I. INTRODUCTION

Initial alignment for inertial navigation system (INS) has been a classical and gradually reborn research topic. It has been widely recognized that the coarse alignment is indispensable for the following extended Kalman filter (EKF) based fine alignment [1]. In this respect, much effort has been focused on the fast coarse alignment methods investigation with optimization-based approach as the prominent representative [2-6]. The necessity of coarse alignment for EKF-based fine alignment lies in making valid the first-order linearization approximation of the navigation parameters error state model. That is to say, this first-order linear error state model applies only to situations with small initial misalignment. This is the truth until the initial alignment meets the invariant EKF (IEKF). In IEKF, through representing the navigation parameters as a matrix Lie group, the corresponding process dynamics possess a special group-affine property [7-11]. Moreover, the vector-form error state on the associated Lie algebra satisfies a trajectory independent log-linear differential equation.

[1] This work was supported in part by the National Natural Science Foundation of China (61873275).
Lubin Chang is with the College of Electrical Engineering, Naval University of Engineering, Wuhan, China. Yarong Luo is with the GNSS Research Center, Wuhan University, Wuhan, China (changlubin@163.com, yarongluo@whu.edu.cn).

Due to such striking property, the IEKF is shown to converge around any trajectory. For INS initial alignment, this means that with the log-linear differential equation, the IEKF is applicable even with arbitrary large misalignments. This has been demonstrated with successful applications and promising results in [8, 9, 12-16].

In [7], Barrau and Bonnabel have provided the rigorous proof of the strong convergence of the log-linear differential equation in continuous time. However, it is undeniable that the log-linear differential equation is still derived with first-order approximation of the attitude error matrix in existing literatures [7, 11-18]. The promising results by IEKF with arbitrary large misalignments seemingly go against the common sense that the first-order linearization should be therefore only valid for small misalignments. Although the uncertainty propagation in discrete time for left invariant error is given in literature [19], it aims at providing covariance propagation for preintegration theory on matrix Lie group $SE_2(3)$ and does not clarify the relationship between the mechanization procedure in inertial navigation system and the uncertainty propagation in preintegration theory. With this consideration, this paper aims to derive the trajectory independent log-linear differential equation in continuous time without any approximation. From the perspective of invariant EKF, our method is novel and clear for INS and can be seen as a nice connection of uncertainty propagation between the discrete time and the continuous time. In fact, the state transition matrices in the discrete time for both left invariant error and right invariant error can be obtained from the error dynamic matrices in the continuous time immediately as the error dynamic matrices can be treated as constants during the time interval [11]. Besides the rigorous proof with "control language" in [7], this work provides another evidence for the validity of the log-linear differential equation with arbitrary attitude misalignments using "navigation language". The definition of the flow in the continuous time which plays a key role in the proof of the log-linear property is inexplicit in literature [7]. However, the analytical and concrete form of the flow is leveraged in our method.

The remaining content is organized as follows. Section II formulates the group affine kinematic model of INS. In this section, the log-linear differential equation is also derived making use of the first-order approximation of the attitude error matrix. Through a group state decomposition procedure, the log-linear differential equation is derived without any approximation in Section III. Finally, Section IV concludes this article.

## II. PROBLEM FORMULATION

Denote by $e$ the Earth-Centered-Earth-Fixed (ECEF) frame, by $b$ the INS body frame, by $i$ the Earth-Centered-Inertial (ECI) frame. The attitude differential equation in earth frame is given by

$$\dot{\mathbf{C}}_b^e = \mathbf{C}_b^e(\boldsymbol{\omega}_{ib}^b \times) - (\boldsymbol{\omega}_{ie}^e \times)\mathbf{C}_b^e \tag{1}$$

where $\boldsymbol{\omega}_{ib}^b$ is the body angular rate measured by gyroscopes in the body frame. $\boldsymbol{\omega}_{ie}^e$ is the earth rotation vector expressed in the earth frame.

The differential equation of the velocity is given by

$$\dot{\mathbf{v}}_{ib}^e = \mathbf{C}_b^e \mathbf{f}^b - (\boldsymbol{\omega}_{ie}^e \times) \mathbf{v}_{ib}^e + \mathbf{G}_{ib}^e \tag{2}$$

where $\mathbf{G}_{ib}^e$ is the gravitational vector expressed in the earth frame and its relationship with gravity vector expressed the earth frame is given by

$$\mathbf{G}_{ib}^e = \mathbf{g}_{ib}^e + (\boldsymbol{\omega}_{ie}^e \times)^2 \mathbf{r}_{ib}^e = \mathbf{g}_{ib}^e + (\boldsymbol{\omega}_{ie}^e \times)^2 \mathbf{r}_{eb}^e \tag{3}$$

where $\mathbf{r}_{eb}^e$ is the position vector in the earth frame. It should be noted that the velocity $\mathbf{v}_{ib}^e$ denotes the body velocity corresponding to the inertial frame expressed in the earth frame. It is not the usual ground velocity $\mathbf{v}_{eb}^e$. The relationship between $\mathbf{v}_{ib}^e$ and $\mathbf{v}_{eb}^e$ is given by

$$\mathbf{v}_{ib}^e = \mathbf{v}_{eb}^e + (\boldsymbol{\omega}_{ie}^e \times) \mathbf{r}_{ib}^e = \mathbf{v}_{eb}^e + (\boldsymbol{\omega}_{ie}^e \times) \mathbf{r}_{eb}^e \tag{4}$$

The velocity provided by the GPS is virtually $\mathbf{v}_{eb}^e$ and the always referred velocity of the vehicle is also $\mathbf{v}_{eb}^e$ and not $\mathbf{v}_{ib}^e$. The reason for the usage of $\mathbf{v}_{ib}^e$ here is that the resulting group state model is group affine, which will be shown in (7) [13, 18, 19, 20].

The differential equation of the position $\mathbf{r}_{ib}^e$ is given by

$$\dot{\mathbf{r}}_{ib}^e = \mathbf{v}_{ib}^e - (\boldsymbol{\omega}_{ie}^e \times) \mathbf{r}_{ib}^e \tag{5}$$

It can be seen from (1) and (2) that the position error will not affect the attitude and velocity error propagation if we ignore the error in gravitational vector $\mathbf{G}_{ib}^e$. Moreover, the attitude $\mathbf{C}_b^e$, velocity $\mathbf{v}_{ib}^e$ and position $\mathbf{r}_{ib}^e$ are assembled as group element of $SE_2(3)$, that is

$$\boldsymbol{\chi} = \begin{bmatrix} \mathbf{C}_b^e & \mathbf{v}_{ib}^e & \mathbf{r}_{ib}^e \\ \mathbf{0}_{1\times 3} & 1 & 0 \\ \mathbf{0}_{1\times 3} & 0 & 1 \end{bmatrix} \in SE_2(3) \tag{6}$$

According to (1), (2) and (5), the differential equation of is given by

$$\dot{\boldsymbol{\chi}} = f_u(\boldsymbol{\chi}) = \begin{bmatrix} \mathbf{C}_b^e(\boldsymbol{\omega}_{ib}^b \times) - (\boldsymbol{\omega}_{ie}^e \times)\mathbf{C}_b^e & \mathbf{C}_b^e \mathbf{f}^b - (\boldsymbol{\omega}_{ie}^e \times)\mathbf{v}_{ib}^e + \mathbf{G}_{ib}^e & -(\boldsymbol{\omega}_{ie}^e \times)\mathbf{r}_{ib}^e + \mathbf{v}_{ib}^e \\ \mathbf{0}_{1\times 3} & 0 & 0 \\ \mathbf{0}_{1\times 3} & 0 & 0 \end{bmatrix}$$

$$= \begin{bmatrix} \mathbf{C}_b^e & \mathbf{v}_{ib}^e & \mathbf{r}_{ib}^e \\ \mathbf{0}_{1\times 3} & 1 & 0 \\ \mathbf{0}_{1\times 3} & 0 & 1 \end{bmatrix} \begin{bmatrix} (\boldsymbol{\omega}_{ib}^b \times) & \mathbf{f}^b & \mathbf{0}_{3\times 1} \\ \mathbf{0}_{1\times 3} & 0 & 0 \\ \mathbf{0}_{1\times 3} & 0 & 0 \end{bmatrix} + \begin{bmatrix} -(\boldsymbol{\omega}_{ie}^e \times) & \mathbf{G}_{ib}^e & \mathbf{0}_{3\times 1} \\ \mathbf{0}_{1\times 3} & 0 & 0 \\ \mathbf{0}_{1\times 3} & 0 & 0 \end{bmatrix} \begin{bmatrix} \mathbf{C}_b^e & \mathbf{v}_{ib}^e & \mathbf{r}_{ib}^e \\ \mathbf{0}_{1\times 3} & 1 & 0 \\ \mathbf{0}_{1\times 3} & 0 & 1 \end{bmatrix} + \begin{bmatrix} \mathbf{0}_{3\times 3} & \mathbf{0}_{3\times 1} & \mathbf{v}_{ib}^e \\ \mathbf{0}_{1\times 3} & 0 & 0 \\ \mathbf{0}_{1\times 3} & 0 & 0 \end{bmatrix} \tag{7}$$

$$= \boldsymbol{\chi}\mathbf{U} + \mathbf{W}\boldsymbol{\chi} + f(\boldsymbol{\chi})$$

where $f$ is a invariant vector field defined by

$$f:\chi = \begin{bmatrix} \mathbf{C} & \mathbf{v} & \mathbf{r} \\ \mathbf{0}_{1\times 3} & 1 & 0 \\ \mathbf{0}_{1\times 3} & 0 & 1 \end{bmatrix} \mapsto \begin{bmatrix} \mathbf{0}_{3\times 3} & \mathbf{0}_{3\times 1} & \mathbf{v} \\ \mathbf{0}_{1\times 3} & 0 & 0 \\ \mathbf{0}_{1\times 3} & 0 & 0 \end{bmatrix} = f(\chi) \tag{8}$$

It is shown that (7) is neither left nor right invariant nor a combination of both and therefore, the system is not invariant. However, the system is group affine with the model satisfying the following condition:

$$f_u(\chi_1\chi_2) = f_u(\chi_1)\chi_2 + \chi_1 f_u(\chi_2) - \chi_1 f_u(\mathbf{I}_d)\chi_2 \tag{9}$$

where $\mathbf{I}_d$ is the group identity element with attitude matrix be identity matrix and three-dimensional vectors be zero vectors. Meanwhile, $f(\cdot)$ is supposed to be group affine, too. As $f(\mathbf{I}_5) = \mathbf{0}_5$, we can obtain that

$$\begin{aligned} f(\chi_1\chi_2) &= f(\chi_1)\chi_2 + \chi_1 f(\chi_2) - \chi_1 f(\mathbf{I}_5)\chi_2 \\ &= f(\chi_1)\chi_2 + \chi_1 f(\chi_2) \end{aligned} \tag{10}$$

Define the left-invariant group error as

$$\boldsymbol{\eta}_l = \chi^{-1}\hat{\chi} = \begin{bmatrix} \mathbf{dC}_l & \mathbf{dv}_l & \mathbf{dr}_l \\ \mathbf{0}_{1\times 3} & 1 & 0 \\ \mathbf{0}_{1\times 3} & 0 & 1 \end{bmatrix} \tag{11}$$

The corresponding trajectory-independent group error model is given by

$$\dot{\boldsymbol{\eta}}_l = g_u(\boldsymbol{\eta}_l) = f_u(\boldsymbol{\eta}_l) - f_u(\mathbf{I}_d)\boldsymbol{\eta}_l \tag{12}$$

Substituting the definition of the left-invariant group error into (12) gives

$$\dot{\boldsymbol{\eta}}_l = \begin{bmatrix} \mathbf{dC}_l(\boldsymbol{\omega}_{ib}^b\times) - (\boldsymbol{\omega}_{ib}^b\times)\mathbf{dC}_l & (\mathbf{dC}_l - \mathbf{I}_{3\times 3})\mathbf{f}^b - (\boldsymbol{\omega}_{ib}^b\times)\mathbf{dv}_l & \mathbf{dv}_l - (\boldsymbol{\omega}_{ib}^b\times)\mathbf{dr}_l \\ \mathbf{0}_{1\times 3} & 0 & 0 \\ \mathbf{0}_{1\times 3} & 0 & 0 \end{bmatrix} \tag{13}$$

Making use of the first-order approximation for the exponential map of SO(3), $\mathbf{dC}_l$ can be approximated as

$$\mathbf{dC}_l \approx \mathbf{I}_{3\times 3} + (\boldsymbol{\varphi}_l\times) \tag{14}$$

Substituting (14) into (13) gives

$$\dot{\boldsymbol{\eta}}_l \approx \begin{bmatrix} -(\boldsymbol{\omega}_{ib}^b\times\boldsymbol{\varphi}_l)\times & (\boldsymbol{\varphi}_l\times)\mathbf{f}^b - (\boldsymbol{\omega}_{ib}^b\times)\mathbf{dv}_l & \mathbf{dv}_l - (\boldsymbol{\omega}_{ib}^b\times)\mathbf{dr}_l \\ \mathbf{0}_{1\times 3} & 0 & 0 \\ \mathbf{0}_{1\times 3} & 0 & 0 \end{bmatrix}$$

$$\tag{15}$$

Define the vector form error corresponding to (11) as $\delta \mathbf{x}_l = \begin{bmatrix} \boldsymbol{\varphi}_l^T & \delta \mathbf{v}_l^T & \delta \mathbf{r}_l^T \end{bmatrix}^T$, where $\mathbf{dv}_l = \mathbf{J}\delta\mathbf{v}_l$, $\mathbf{dr}_l = \mathbf{J}\delta\mathbf{r}_l$, $\mathbf{J}$ is the left Jacobian matrix of $\boldsymbol{\varphi}_l$. Then, the corresponding model can be deduced from (15), that is

$$\delta \dot{\mathbf{x}}_l = \begin{bmatrix} -(\boldsymbol{\omega}_{ib}^b \times) & \mathbf{0}_{3\times 3} & \mathbf{0}_{3\times 3} \\ -(\mathbf{f}^b \times) & -(\boldsymbol{\omega}_{ib}^b \times) & \mathbf{0}_{3\times 3} \\ \mathbf{0}_{3\times 3} & \mathbf{I}_{3\times 3} & -(\boldsymbol{\omega}_{ib}^b \times) \end{bmatrix} \delta \mathbf{x}_l = \mathbf{F}_l \delta \mathbf{x}_l \tag{16}$$

Similarly, define the right-invariant group error as

$$\boldsymbol{\eta}_r = \hat{\boldsymbol{\chi}} \boldsymbol{\chi}^{-1} = \begin{bmatrix} \mathbf{dC}_r & \mathbf{dv}_r & \mathbf{dr}_r \\ \mathbf{0}_{1\times 3} & 1 & 0 \\ \mathbf{0}_{1\times 3} & 0 & 1 \end{bmatrix} \tag{17}$$

The corresponding trajectory-independent group error model is given by

$$\dot{\boldsymbol{\eta}}_r = g_u(\boldsymbol{\eta}_r) = f_u(\boldsymbol{\eta}_r) - \boldsymbol{\eta}_r f_u(\mathbf{I}_d) \tag{18}$$

Substituting the definition of the right-invariant group error into (18) gives

$$\dot{\boldsymbol{\eta}}_r = \begin{bmatrix} \mathbf{dC}_r(\boldsymbol{\omega}_{ie}^e \times) - (\boldsymbol{\omega}_{ie}^e \times)\mathbf{dC}_r & (\mathbf{I}_{3\times 3} - \mathbf{dC}_r)\mathbf{G}_{ib}^e - (\boldsymbol{\omega}_{ie}^e \times)\mathbf{dv}_r & \mathbf{dv}_r - (\boldsymbol{\omega}_{ie}^e \times)\mathbf{dr}_r \\ \mathbf{0}_{1\times 3} & 0 & 0 \\ \mathbf{0}_{1\times 3} & 0 & 0 \end{bmatrix} \tag{19}$$

Making use of the first-order approximation for the exponential map of SO(3), $\mathbf{dC}_r$ can be approximated as

$$\mathbf{dC}_r \approx \mathbf{I}_{3\times 3} + (\boldsymbol{\varphi}_r \times) \tag{20}$$

Substituting (20) into (19) gives

$$\dot{\boldsymbol{\eta}}_r \approx \begin{bmatrix} -(\boldsymbol{\omega}_{ie}^e \times \boldsymbol{\varphi}_r) \times & (\mathbf{G}_{ib}^e \times)\boldsymbol{\varphi}_r - (\boldsymbol{\omega}_{ie}^e \times)\mathbf{dv}_r & \mathbf{dv}_r - (\boldsymbol{\omega}_{ie}^e \times)\mathbf{dr}_r \\ \mathbf{0}_{1\times 3} & 0 & 0 \\ \mathbf{0}_{1\times 3} & 0 & 0 \end{bmatrix} \tag{21}$$

Define the vector form error corresponding to (17) as $\delta \mathbf{x}_r = \begin{bmatrix} \boldsymbol{\varphi}_r^T & \delta \mathbf{v}_r^T & \delta \mathbf{r}_r^T \end{bmatrix}^T$, where $\mathbf{dv}_r = \mathbf{J}\delta\mathbf{v}_r$, $\mathbf{dr}_r = \mathbf{J}\delta\mathbf{r}_r$, $\mathbf{J}$ is the left Jacobian matrix of $\boldsymbol{\varphi}_r$. Then, the corresponding model can be deduced from (21), that is

$$\delta \dot{\mathbf{x}}_r = \begin{bmatrix} -(\boldsymbol{\omega}_{ie}^e \times) & \mathbf{0}_{3\times 3} & \mathbf{0}_{3\times 3} \\ (\mathbf{G}_{ib}^e \times) & -(\boldsymbol{\omega}_{ie}^e \times) & \mathbf{0}_{3\times 3} \\ \mathbf{0}_{3\times 3} & \mathbf{I}_{3\times 3} & -(\boldsymbol{\omega}_{ie}^e \times) \end{bmatrix} \delta \mathbf{x}_r = \mathbf{F}_r \delta \mathbf{x}_r \tag{22}$$

It is shown that the trajectory-independent vector error equations (16) and (22) are derived based on the first-order approximation for the exponential map of SO(3). It is known that such approximation or linearization is only valid when the attitude error is small. However, the models (16) and (22) can be used to accomplish initial alignment with arbitrary large misalignment, which seemingly goes against the aforementioned approximation. In the following section, we will demonstrate that the models (16) and (22) can also be derived without any approximation.

## III. VECTOR ERROR MODELS DERIVATION WITHOUT APPROXIMATION

### A. GROUP STATE MODEL RECONSTRUCTION

In order to derive the models (16) and (22) without approximation, the group state model (7) is firstly reconstructed through group state decomposition. In recent few years, the concept of attitude matrix composition has been used in attitude determination or estimation based initial alignment [2-6] and invariant EKF [11]. Denote the initial inertially fixed body frame as $b_0$, the initial inertially fixed earth frame as $e_0$, the attitude matrix $\mathbf{C}_b^e$ can be decomposed as

$$\mathbf{C}_b^e = \mathbf{C}_{e_0}^e \mathbf{C}_{b_0}^{e_0} \mathbf{C}_b^{b_0} \tag{23}$$

Similar to the decomposition of attitude matrix, the navigation state which is embedded into matrix Lie group can also be decomposed as [19, 21, 22, 24]

$$\chi = \chi_e \Phi_t(\chi_0) \chi_b \tag{24}$$

where

$$\chi_0 = \begin{bmatrix} \mathbf{C}_{b_0}^{e_0} & \mathbf{v}_{ib,0}^e & \mathbf{r}_{ib,0}^e \\ \mathbf{0}_{1\times 3} & 1 & 0 \\ \mathbf{0}_{1\times 3} & 0 & 1 \end{bmatrix} \tag{25a}$$

$$\chi_e = \begin{bmatrix} \mathbf{C}_{e_0}^e & \bar{\mathbf{v}}_{ib}^e & \bar{\mathbf{r}}_{ib}^e \\ \mathbf{0}_{1\times 3} & 1 & 0 \\ \mathbf{0}_{1\times 3} & 0 & 1 \end{bmatrix} \tag{25b}$$

$$\chi_b = \begin{bmatrix} \mathbf{C}_b^{b_0} & \mathbf{v}_{ib}^{b_0} & \mathbf{r}_{ib}^{b_0} \\ \mathbf{0}_{1\times 3} & 1 & 0 \\ \mathbf{0}_{1\times 3} & 0 & 1 \end{bmatrix} \tag{25c}$$

$$\Phi_t : \chi = \begin{bmatrix} \mathbf{C} & \mathbf{v} & \mathbf{r} \\ \mathbf{0}_{1\times 3} & 1 & 0 \\ \mathbf{0}_{1\times 3} & 0 & 1 \end{bmatrix} \mapsto \begin{bmatrix} \mathbf{C} & \mathbf{v} & \mathbf{r}+t\mathbf{v} \\ \mathbf{0}_{1\times 3} & 1 & 0 \\ \mathbf{0}_{1\times 3} & 0 & 1 \end{bmatrix} = \chi + tf(\chi) \tag{25d}$$

$\Phi_t(\cdot)$ can be seen as the concrete definition of flow in the continuous time for navigation state. It is worth noting that the position part is related with the velocity which is consistent with our intuitive. It is easy to verify that $\Phi_t(\cdot)$ is a group automorphism and owns the log-linearity property and satisfies [19]

$$\Phi_t(\exp_m(\xi^\wedge)) = \exp_m((\mathbf{F}\xi)^\wedge),$$

$$\mathbf{F} := \mathbf{F}_t = \begin{bmatrix} \mathbf{I}_{3\times3} & \mathbf{0}_{3\times3} & \mathbf{0}_{3\times3} \\ \mathbf{0}_{3\times3} & \mathbf{I}_{3\times3} & \mathbf{0}_{3\times3} \\ \mathbf{0}_{3\times3} & t\mathbf{I}_{3\times3} & \mathbf{I}_{3\times3} \end{bmatrix} \tag{26}$$

where the operator $(\cdot)^\wedge$ is used to map the vector in Euclidean space to elements of Lie algebra. Combining equation (8) and equation (25d), we can obtain that $f(\Phi_t(\chi)) = f(\chi)$, which is will be used for many times.

It is shown that $\chi_0$ is the initial extended pose and is constant. $\chi_b$ is a local increment and the kinematic equation of $\chi_b$ is given by

$$\dot{\chi}_b = \begin{bmatrix} \mathbf{C}_b^{b_0}(\boldsymbol{\omega}_{ib}^b\times) & \mathbf{C}_b^{b_0}\mathbf{f}^b & \mathbf{v}_{ib}^{b_0} \\ \mathbf{0}_{1\times3} & 0 & 0 \\ \mathbf{0}_{1\times3} & 0 & 0 \end{bmatrix} = \chi_b \mathbf{U} + f(\chi_b) \tag{27}$$

where $\mathbf{U}$ is defined in (7). It can be easily obtained that the initial value of $\chi_b$ is $\mathbf{I}_5$, that is error-free. Similarly, $\chi_e$ is a global increment and the kinematic equation of $\chi_e$ is given by

$$\dot{\chi}_e = \begin{bmatrix} -(\boldsymbol{\omega}_{ie}^e\times)\mathbf{C}_{e_0}^e & -(\boldsymbol{\omega}_{ie}^e\times)\overline{\mathbf{v}}_{ib}^e + \mathbf{G}_{ib}^e & \overline{\mathbf{v}}_{ib}^e - (\boldsymbol{\omega}_{ie}^e\times)\overline{\mathbf{r}}_{ib}^e \\ \mathbf{0}_{1\times3} & 0 & 0 \\ \mathbf{0}_{1\times3} & 0 & 0 \end{bmatrix} = \mathbf{W}\chi_e + f(\chi_e) \tag{28}$$

where $\mathbf{W}$ is defined in (7). It can also be easily obtained that the initial value of $\chi_e$ is $\mathbf{I}_5$, that is also error-free. It should be noted that the group state decomposition is also the key point for accurate preintegration in inertial based robotics applications [19, 21-24]. The preintegration is another striking property of the group affine system besides the trajectory-independent invariant error propagation.

With the initial navigation state $(\mathbf{C}_{b_0}^{e_0}, \mathbf{v}_{ib,0}^e, \mathbf{r}_{ib,0}^e)$, the navigation state defined in equation (24) can be written separately as

$$\mathbf{C}_b^e = \mathbf{C}_{e_0}^e \mathbf{C}_{b_0}^{e_0} \mathbf{C}_b^{b_0} \tag{29}$$

$$\mathbf{v}_{ib}^e = \overline{\mathbf{v}}_{ib}^e + \mathbf{C}_{e_0}^e (\mathbf{C}_{b_0}^{e_0} \mathbf{v}_{ib}^{b_0} + \mathbf{v}_{ib,0}^e) \tag{30}$$

$$\mathbf{r}_{ib}^{e} = \overline{\mathbf{r}}_{ib}^{e} + \mathbf{C}_{e_0}^{e} (\mathbf{C}_{b_0}^{e_0} \mathbf{r}_{ib}^{b_0} + \mathbf{r}_{ib,0}^{e} + t\mathbf{v}_{ib,0}^{e}) \tag{31}$$

Since the initial value and inputs of model (28) are both error-free, $\chi_e$ will not introduce error in $\chi$. In other words, the error of $\chi$ only comes from $\chi_0$ (initial error) and $\chi_b$ (measurement error of the inertial sensors). In order of facilitate the following derivation, we can combine $\chi_0$ and $\chi_b$ as

$$\tilde{\chi}_b = \Phi_t(\chi_0)\chi_b = \begin{bmatrix} \mathbf{C}_b^{e_0} & \overline{\mathbf{v}}_{ib}^{e_0} & \overline{\mathbf{r}}_{ib}^{e_0} \\ \mathbf{0}_{1\times 3} & 1 & 0 \\ \mathbf{0}_{1\times 3} & 0 & 1 \end{bmatrix} \tag{32}$$

Combining $\chi_0$ and $\chi_b$ is similar with the procedure that combines $\mathbf{C}_{b_0}^{e_0}$ and $\mathbf{C}_b^{b_0}$ as in [6]. From (25a) and (25c), it can be known that

$$\overline{\mathbf{v}}_{ib}^{e_0} = \mathbf{C}_{b_0}^{e_0} \mathbf{v}_{ib}^{b_0} + \mathbf{v}_{ib,0}^{e} \tag{33}$$

$$\overline{\mathbf{r}}_{ib}^{e_0} = \mathbf{C}_{b_0}^{e_0} \mathbf{r}_{ib}^{b_0} + \mathbf{r}_{ib,0}^{e} + t\mathbf{v}_{ib,0}^{e} \tag{34}$$

The kinematic equation of $\tilde{\chi}_b$ is given by

$$\dot{\tilde{\chi}}_b = \begin{bmatrix} \mathbf{C}_b^{e_0}(\boldsymbol{\omega}_{ib}^{b}\times) & \mathbf{C}_b^{e_0}\mathbf{f}^b & \overline{\mathbf{v}}_{ib}^{e_0} \\ \mathbf{0}_{1\times 3} & 0 & 0 \\ \mathbf{0}_{1\times 3} & 0 & 0 \end{bmatrix} = \tilde{\chi}_b \mathbf{U} + f(\tilde{\chi}_b) \tag{35}$$

which has a same form as (27). However, different with $\chi_b$ that has an error-free initial values, the initial value of $\tilde{\chi}_b$ is $\chi_0$, which is unknown and may be far from the true value. The unknown value $\chi_0$ is also the reason for large misalignment in initial alignment.

### B. VECTOR FORM LEFT-ERROR MODEL DERIVATION WITHOUT APPROXIMATION

According to the group state decomposition (24), the left-invariant group error in (11) can be re-derived as

$$\boldsymbol{\eta}_l = \chi^{-1}\hat{\chi} = \tilde{\chi}_b^{-1}\hat{\tilde{\chi}}_b \tag{36}$$

Moreover, substituting equation (36) into equation (12), the left-invariant group error model (13) can be reorganized as

$$\dot{\boldsymbol{\eta}}_l = g_u(\boldsymbol{\eta}_l) = \boldsymbol{\eta}_l \mathbf{U} - \mathbf{U}\boldsymbol{\eta}_l + f(\boldsymbol{\eta}_l) \tag{37}$$

Let $\boldsymbol{\eta}_{l,0}$ be the initial left-invariant group error and its associated Lie algebra can be denoted by the vector $\delta\mathbf{x}_{l,0}$. The matrix Lie group is related to its associated vector form in Euclidean space through the exponential map, that is

$$\boldsymbol{\eta}_{l,0} = \exp(\delta\mathbf{x}_{l,0}) = \exp_m(\delta\mathbf{x}_{l,0}^{\wedge}) \tag{38}$$

Define the following group error

$$\boldsymbol{\eta}_l = \tilde{\boldsymbol{\chi}}_b^{-1} \Phi_t(\boldsymbol{\eta}_{l,0}) \tilde{\boldsymbol{\chi}}_b \tag{39}$$

The decomposition of the error state can be thought of as shifting the navigation state inside the exponential mapping using the group Adjoint operator definition, then the log-linear error dynamics can be recovered. The decomposition is similar to the rotation matrix decomposition in [11], and we believe that it is reasonable. Moreover, the subsequent results do confirm that the decomposition is correct.

Taking derivative of both sides of (39) gives

$$\dot{\boldsymbol{\eta}}_l = -\tilde{\boldsymbol{\chi}}_b^{-1} \dot{\tilde{\boldsymbol{\chi}}}_b \tilde{\boldsymbol{\chi}}_b^{-1} \Phi_t(\boldsymbol{\eta}_{l,0}) \tilde{\boldsymbol{\chi}}_b + \tilde{\boldsymbol{\chi}}_b^{-1} \dot{\Phi}_t(\boldsymbol{\eta}_{l,0}) \tilde{\boldsymbol{\chi}}_b + \tilde{\boldsymbol{\chi}}_b^{-1} \Phi_t(\boldsymbol{\eta}_{l,0}) \dot{\tilde{\boldsymbol{\chi}}}_b \tag{40}$$

Substituting (35) into (40) and using equation (25d) we can get

$$\begin{aligned}\dot{\boldsymbol{\eta}}_l &= -\tilde{\boldsymbol{\chi}}_b^{-1} \tilde{\boldsymbol{\chi}}_b \mathbf{U} \tilde{\boldsymbol{\chi}}_b^{-1} \Phi_t(\boldsymbol{\eta}_{l,0}) \tilde{\boldsymbol{\chi}}_b + \tilde{\boldsymbol{\chi}}_b^{-1} f(\boldsymbol{\eta}_{l,0}) \tilde{\boldsymbol{\chi}}_b + \tilde{\boldsymbol{\chi}}_b^{-1} \Phi_t(\boldsymbol{\eta}_{l,0}) \tilde{\boldsymbol{\chi}}_b \mathbf{U} \\ &\quad - \tilde{\boldsymbol{\chi}}_b^{-1} \tilde{\boldsymbol{\chi}}_b f(\tilde{\boldsymbol{\chi}}_b) \tilde{\boldsymbol{\chi}}_b^{-1} \Phi_t(\boldsymbol{\eta}_{l,0}) \tilde{\boldsymbol{\chi}}_b + \tilde{\boldsymbol{\chi}}_b^{-1} \Phi_t(\boldsymbol{\eta}_{l,0}) f(\tilde{\boldsymbol{\chi}}_b) \\ &= -\mathbf{U} \boldsymbol{\eta}_l + \boldsymbol{\eta}_l \mathbf{U} + f(\boldsymbol{\eta}_l)\end{aligned} \tag{41}$$

which means that the group error defined in (39) is also a solution to the group error dynamics equation (37). In the derivation, a nice property of function defined in equation (8) is used, that is $f(\Phi_t(\boldsymbol{\eta}_{l,0})) = f(\boldsymbol{\eta}_{l,0})$. The detailed derivation is omitted in order to save space.

According to the relationship between the matrix Lie group and its associated Lie algebra, (39) can be rewritten as

$$\begin{aligned}\exp(\boldsymbol{\delta x}_l) &= \tilde{\boldsymbol{\chi}}_b^{-1} \Phi_t(\exp(\boldsymbol{\delta x}_{l,0})) \tilde{\boldsymbol{\chi}}_b \\ &= \tilde{\boldsymbol{\chi}}_b^{-1} (\exp(\mathbf{F} \boldsymbol{\delta x}_{l,0})) \tilde{\boldsymbol{\chi}}_b = \exp(\mathrm{Ad}_{\tilde{\boldsymbol{\chi}}_b^{-1}} \mathbf{F} \boldsymbol{\delta x}_{l,0})\end{aligned} \tag{42}$$

where $\mathrm{Ad}_{\tilde{\boldsymbol{\chi}}_b^{-1}}$ is the Adjoint operator of $\tilde{\boldsymbol{\chi}}_b^{-1}$ and is given by

$$\mathrm{Ad}_{\tilde{\boldsymbol{\chi}}_b^{-1}} = \begin{bmatrix} \mathbf{C}_b^{e_0 T} & \mathbf{0}_{3\times 3} & \mathbf{0}_{3\times 3} \\ -\mathbf{C}_b^{e_0 T} (\bar{\mathbf{v}}_{ib}^{e_0} \times) & \mathbf{C}_b^{e_0 T} & \mathbf{0}_{3\times 3} \\ -\mathbf{C}_b^{e_0 T} (\bar{\mathbf{r}}_{ib}^{e_0} \times) & \mathbf{0}_{3\times 3} & \mathbf{C}_b^{e_0 T} \end{bmatrix} \tag{43}$$

According to (42) we can get the following equation

$$\boldsymbol{\delta x}_l = \mathrm{Ad}_{\tilde{\boldsymbol{\chi}}_b^{-1}} \mathbf{F} \boldsymbol{\delta x}_{l,0} \tag{44}$$

Taking derivative of both sides of (44) gives

$$\delta \dot{\mathbf{x}}_l = \begin{bmatrix} -(\boldsymbol{\omega}_{ib}^b \times) \mathbf{C}_b^{e_0 T} & \mathbf{0}_{3\times 3} & \mathbf{0}_{3\times 3} \\ (\boldsymbol{\omega}_{ib}^b \times) \mathbf{C}_b^{e_0 T} (\overline{\mathbf{v}}_{ib}^{e_0} \times) - (\mathbf{f}^b \times) \mathbf{C}_b^{e_0 T} & -(\boldsymbol{\omega}_{ib}^b \times) \mathbf{C}_b^{e_0 T} & \mathbf{0}_{3\times 3} \\ (\boldsymbol{\omega}_{ib}^b \times) \mathbf{C}_b^{e_0 T} (\overline{\mathbf{r}}_{ib}^{e_0} \times) - \mathbf{C}_b^{e_0 T} (\overline{\mathbf{v}}_{ib}^{e_0} \times) & \mathbf{C}_b^{e_0 T} - t(\boldsymbol{\omega}_{ib}^b \times) \mathbf{C}_b^{e_0 T} & -(\boldsymbol{\omega}_{ib}^b \times) \mathbf{C}_b^{e_0 T} \end{bmatrix} \delta \mathbf{x}_{l,0}$$

$$= \begin{bmatrix} -(\boldsymbol{\omega}_{ib}^b \times) & \mathbf{0}_{3\times 3} & \mathbf{0}_{3\times 3} \\ -(\mathbf{f}^b \times) & -(\boldsymbol{\omega}_{ib}^b \times) & \mathbf{0}_{3\times 3} \\ \mathbf{0}_{3\times 3} & \mathbf{I}_{3\times 3} & -(\boldsymbol{\omega}_{ib}^b \times) \end{bmatrix} \begin{bmatrix} \mathbf{C}_b^{e_0 T} & \mathbf{0}_{3\times 3} & \mathbf{0}_{3\times 3} \\ -\mathbf{C}_b^{e_0 T} (\overline{\mathbf{v}}_{ib}^{e_0} \times) & \mathbf{C}_b^{e_0 T} & \mathbf{0}_{3\times 3} \\ -\mathbf{C}_b^{e_0 T} (\overline{\mathbf{r}}_{ib}^{e_0} \times) & \mathbf{0}_{3\times 3} & \mathbf{C}_b^{e_0 T} \end{bmatrix} \begin{bmatrix} \mathbf{I}_{3\times 3} & \mathbf{0}_{3\times 3} & \mathbf{0}_{3\times 3} \\ \mathbf{0}_{3\times 3} & \mathbf{I}_{3\times 3} & \mathbf{0}_{3\times 3} \\ \mathbf{0}_{3\times 3} & t\mathbf{I}_{3\times 3} & \mathbf{I}_{3\times 3} \end{bmatrix} \delta \mathbf{x}_{l,0} \quad (45)$$

$$= \mathbf{F}_l \, \mathrm{Ad}_{\tilde{\chi}_b^{-1}} \mathbf{F} \delta \mathbf{x}_{l,0} = \mathbf{F}_l \delta \mathbf{x}_l$$

which is just the model in (16). It is shown that there is no approximation from (38) to (45). That is to say, the trajectory-independent vector error state model (16) can be derived without any approximation. Now, this can be understandable that why the error state model derived through "first-order linearization" can be applied in initial alignment with arbitrary large misalignments.

## C. VECTOR FORM RIGHT-ERROR MODEL DERIVATION WITHOUT APPROXIMATION

Next, we make use of similar procedure to derive the vector form right-error model (22) without any approximation.

The right-invariant group error model (19) can be reorganized as

$$\dot{\boldsymbol{\eta}}_r = \mathbf{W} \boldsymbol{\eta}_r - \boldsymbol{\eta}_r \mathbf{W} + f(\boldsymbol{\eta}_r) \quad (46)$$

Let $\boldsymbol{\eta}_{r,0}$ be the initial right-invariant group error and its associated Lie algebra can be denoted by the vector $\delta \mathbf{x}_{r,0}$. Similar to (38), we can get

$$\boldsymbol{\eta}_{r,0} = \exp(\delta \mathbf{x}_{r,0}) = \exp_{\mathrm{m}}(\delta \mathbf{x}_{r,0}^{\wedge}) \quad (47)$$

Define the following group error

$$\boldsymbol{\eta}_r = \boldsymbol{\chi}_e \Phi_t(\boldsymbol{\eta}_{r,0}) \boldsymbol{\chi}_e^{-1} \quad (48)$$

Taking derivative of both sides of (48) gives

$$\dot{\boldsymbol{\eta}}_r = \dot{\boldsymbol{\chi}}_e \Phi_t(\boldsymbol{\eta}_{r,0}) \boldsymbol{\chi}_e^{-1} + \boldsymbol{\chi}_e \dot{\Phi}_t(\boldsymbol{\eta}_{r,0}) \boldsymbol{\chi}_e^{-1} - \boldsymbol{\chi}_e \Phi_t(\boldsymbol{\eta}_{r,0}) \boldsymbol{\chi}_e^{-1} \dot{\boldsymbol{\chi}}_e \boldsymbol{\chi}_e^{-1} \quad (49)$$

Substituting (25) into (42) gives

$$\begin{aligned} \dot{\boldsymbol{\eta}}_r &= \mathbf{W} \boldsymbol{\chi}_e \Phi_t(\boldsymbol{\eta}_{r,0}) \boldsymbol{\chi}_e^{-1} + \boldsymbol{\chi}_e f(\boldsymbol{\eta}_{r,0}) \boldsymbol{\chi}_e^{-1} - \boldsymbol{\chi}_e \Phi_t(\boldsymbol{\eta}_{r,0}) \boldsymbol{\chi}_e^{-1} \mathbf{W} \boldsymbol{\chi}_e \boldsymbol{\chi}_e^{-1} \\ &\quad + f(\boldsymbol{\chi}_e) \Phi_t(\boldsymbol{\eta}_{r,0}) \boldsymbol{\chi}_e^{-1} - \boldsymbol{\chi}_e \Phi_t(\boldsymbol{\eta}_{r,0}) \boldsymbol{\chi}_e^{-1} f(\boldsymbol{\chi}_e) \boldsymbol{\chi}_e^{-1} \\ &= \mathbf{W} \boldsymbol{\eta}_r - \boldsymbol{\eta}_r \mathbf{W} + f(\boldsymbol{\eta}_r) \end{aligned} \quad (50)$$

which means that the group error defined in (48) is also a solution to the group error dynamics equation (46).

Similar to (42), (48) can be rewritten as

$$\exp(\delta \mathbf{x}_r) = \chi_e \Phi_t(\exp(\delta \mathbf{x}_{r,0})) \chi_e^{-1} \\ = \chi_e \exp(\mathbf{F} \delta \mathbf{x}_{r,0}) \chi_e^{-1} = \exp(\mathrm{Ad}_{\chi_e} \mathbf{F} \delta \mathbf{x}_{r,0}) \tag{51}$$

where $\mathrm{Ad}_{\chi_e}$ is the Adjoint operator of $\chi_e$ and is given by

$$\mathrm{Ad}_{\chi_e} = \begin{bmatrix} \mathbf{C}_{e_0}^e & \mathbf{0}_{3\times 3} & \mathbf{0}_{3\times 3} \\ (\overline{\mathbf{v}}_{ib}^e \times)\mathbf{C}_{e_0}^e & \mathbf{C}_{e_0}^e & \mathbf{0}_{3\times 3} \\ (\overline{\mathbf{r}}_{ib}^e \times)\mathbf{C}_{e_0}^e & \mathbf{0}_{3\times 3} & \mathbf{C}_{e_0}^e \end{bmatrix} \tag{52}$$

According to (51), we can get

$$\delta \mathbf{x}_r = \mathrm{Ad}_{\chi_e} \mathbf{F} \delta \mathbf{x}_{r,0} \tag{53}$$

Taking derivative of both sides of (53) gives

$$\delta \dot{\mathbf{x}}_r = \begin{bmatrix} -(\boldsymbol{\omega}_{ie}^e \times)\mathbf{C}_{e_0}^e & \mathbf{0}_{3\times 3} & \mathbf{0}_{3\times 3} \\ [(\mathbf{G}_{ib}^e \times) - (\boldsymbol{\omega}_{ie}^e \times)(\overline{\mathbf{v}}_{ib}^e \times)]\mathbf{C}_{e_0}^e & -(\boldsymbol{\omega}_{ie}^e \times)\mathbf{C}_{e_0}^e & \mathbf{0}_{3\times 3} \\ [(\overline{\mathbf{v}}_{ib}^e \times) - (\boldsymbol{\omega}_{ie}^e \times)(\overline{\mathbf{r}}_{ib}^e \times)]\mathbf{C}_{e_0}^e & \mathbf{C}_{e_0}^e - t(\boldsymbol{\omega}_{ie}^e \times)\mathbf{C}_{e_0}^e & -(\boldsymbol{\omega}_{ie}^e \times)\mathbf{C}_{e_0}^e \end{bmatrix} \delta \mathbf{x}_{r,0} \\ = \begin{bmatrix} -(\boldsymbol{\omega}_{ie}^e \times) & \mathbf{0}_{3\times 3} & \mathbf{0}_{3\times 3} \\ (\mathbf{G}_{ib}^e \times) & -(\boldsymbol{\omega}_{ie}^e \times) & \mathbf{0}_{3\times 3} \\ \mathbf{0}_{3\times 3} & \mathbf{I}_{3\times 3} & -(\boldsymbol{\omega}_{ie}^e \times) \end{bmatrix} \begin{bmatrix} \mathbf{C}_{e_0}^e & \mathbf{0}_{3\times 3} & \mathbf{0}_{3\times 3} \\ (\overline{\mathbf{v}}_{ib}^e \times)\mathbf{C}_{e_0}^e & \mathbf{C}_{e_0}^e & \mathbf{0}_{3\times 3} \\ (\overline{\mathbf{r}}_{ib}^e \times)\mathbf{C}_{e_0}^e & \mathbf{0}_{3\times 3} & \mathbf{C}_{e_0}^e \end{bmatrix} \begin{bmatrix} \mathbf{I}_{3\times 3} & \mathbf{0}_{3\times 3} & \mathbf{0}_{3\times 3} \\ \mathbf{0}_{3\times 3} & \mathbf{I}_{3\times 3} & \mathbf{0}_{3\times 3} \\ \mathbf{0}_{3\times 3} & t\mathbf{I}_{3\times 3} & \mathbf{I}_{3\times 3} \end{bmatrix} \delta \mathbf{x}_{r,0} \\ = \mathbf{F}_r \mathrm{Ad}_{\chi_e} \mathbf{F} \delta \mathbf{x}_{r,0} = \mathbf{F}_r \delta \mathbf{x}_r \tag{54}$$

which is just the model in (22). It is shown that there is no approximation from (47) to (54). That is to say, the trajectory-independent vector error state model (22) can be derived without any approximation.

It should be noted that the above derivations are all based on the fact that the error in gravitational vector caused by the position is ignored. If this error is also considered, the velocity error differential equation will have global state components, that is to say, it is not trajectory-independent as shown in [14].

## IV. CONCLUSION

Through matrix Lie group decomposition, this work derives the log-linear error state model for the group affine system without any assumption. The corresponding derivations can provide an explicit explanation why the log-linear error state model can be used to perform INS initial alignment with arbitrary large misalignments.

# REFERENCES

[1] P. D. Groves, Principles of GNSS, Inertial, and Multisensor Integrated Navigation Systems: Artech House, 2008.

[2] P. M. G. Silson, "Coarse Alignment of a Ship's Strapdown Inertial Attitude Reference System Using Velocity Loci," IEEE Transactions on Instrumentation and Measurement, vol. 60, no. 6, pp. 1930-1941, 2011.

[3] M. P. Wu, Y. X. Wu, X. P. Hu and D. W. Hu, "Optimization-based Alignment for Inertial Navigation Systems: Theory and Algorithm," Aerospace Science and Technology, vol. 15, pp. 1-17, 2011.

[4] Y. X. Wu and X. F. Pan, "Velocity/position integration formula, Part I: Application to in-flight coarse alignment," IEEE Transactions on Aerospace and Electronic Systems, vol. 49, no. 2, pp. 1006-1023, 2013.

[5] L. B. Chang, J. S. Li and K. L. Li, "Optimization-based Alignment for Strapdown Inertial Navigation System: Comparison and Extension," IEEE Transactions on Aerospace and Electronic Systems, vol.52, no. 4, pp. 1697-1713, 2016.

[6] L. B. Chang, J. S. Li and S. Y. Chen, "Initial Alignment by Attitude Estimation for Strapdown Inertial Navigation Systems," IEEE Transactions on Instrumentation and Measurement, vol. 64, no. 3, pp. 784-794, 2015.

[7] A. Barrau and S. Bonnabel, "The Invariant Extended Kalman Filter as a Stable Observer," IEEE Transaction on Automatic Control, vol. 62, no. 4, pp. 1797–1812, 2017.

[8] A. Barrau and S. Bonnabel, "Invariant Kalman filtering," Annual Review of Control Robotics and Autonomous Systems, vol. 1, pp. 237–257, May 2018.

[9] A. Barrau, Non-linear state error based extended Kalman filters with applications to navigation. PhD thesis, Mines Paristech, 2015.

[10] A. Barrau and S. Bonnabel, "The Geometry of Navigation Problems," IEEE Transaction on Automatic Control, early access, DOI: 10.1109/TAC.2022.3144328, 2022.

[11] R. Hartley, M. Ghaffari, R. M. Eustice and J. W. Grizzle, "Contact-Aided Invariant Extended Kalman Filtering for Robot State Estimation," The International Journal of Robotics Research, vol. 39, no. 4, pp. 402-430, 2020..

[12] L. B. Chang, F. J. Qin and J. N. Xu, "Strapdown Inertial Navigation System Initial Alignment based on Group of Double Direct Spatial Isometries," IEEE Sensors Journal, vol. 22, no. 1, pp. 803–818, 2022.

[13]   L. B. Chang, J. B. Di and F. J. Qin, "Inertial based Integration with Transformed INS Mechanization in Earth Frame," IEEE/ASME Transactions on Mechatronics, early access, DOI: 10.1109/TMECH.2021.3090428, 2021.

[14]   W. Ouyang and Y. X. Wu, "A Trident Quaternion Framework for Inertial based Navigation Part II: Error Models and Application to Initial Alignment," IEEE Transactions on Aerospace and Electronic Systems, early access, DOI: 10.1109/TAES.2021.3133219, 2021.

[15]   Y. R. Luo, M. Y. Wang, C. Guo and W. F. Guo, "Research on Invariant Extended Kalman Filter Based 5G/SINS Integrated Navigation Simulation," In China Satellite Navigation Conference (CSNC 2021) Proceedings (pp. 455-466). Springer, Singapore.

[16]   Y. R. Luo, C. Guo and J. N. Liu, "Equivariant filtering framework for inertial-integrated navigation," Satellite Navigation, DOI: 10.1186/s43020-021-00061-z, 2021

[17]   M. Brossard, A. Barrau, and S. Bonnabel, "Exploiting Symmetries to Design EKFs With Consistency Properties for Navigation and SLAM," IEEE Sensors Journal, vol. 19, no. 4, pp. 1572–1579, Sep. 2019.

[18]   S. Heo and C. Park, "Consistent EKF-based visual-inertial odometry on matrix Lie group," IEEE Sensors Journal, vol. 18, no. 9, pp. 3780–3788, May 2018.

[19]   M. Brossard, A. Barrau, P. Chauchat and S. Bonnabel, "Associating Uncertainty to Extended Poses for on Lie Group IMU Preintegration with Rotating Earth," IEEE Transaction on Robotics, early access, DOI: 10.1109/TRO.2021.3100156, 2021.

[20]   J. R. Cui, M. S. Wang W. Q. Wu and X. F. He, "Lie group based nonlinear state errors for MEMS-IMU/GNSS/magnetometer integrated navigation," The Journal of Navigation, vol. 74, no. 4, pp. 887-900, 2021.

[21]   A. Barrau and S. Bonnabel, "Linear observed systems on groups," Syst. Control Lett., vol. 129, pp. 36–42, 2019.

[22]   A. Barrau and S. Bonnabel, "A mathematical framework for IMU error propagation with applications to preintegration," in Proc. Int. Conf. Robot. Autom., 2020, pp. 5732–5738.

[23]   C. Forster, L. Carlone, F. Dellaert, and D. Scaramuzza, "On-manifold preintegration for real-time visual-inertial odometry," IEEE Trans. Robot., vol. 33, no. 1, pp. 1–21, Feb. 2017.

[24]   Y. R. Luo, Y. Liu, C. Guo and J. N. Liu, "The Unified Mathematical Framework for IMU Preintegration in Inertial-Aided Navigation System," arXiv:2111.09100v4, 2021.